\documentclass[letterpaper]{article} % DO NOT CHANGE THIS
\usepackage[submission]{aaai23_}  % DO NOT CHANGE THIS
\usepackage{times}  % DO NOT CHANGE THIS
\usepackage{helvet}  % DO NOT CHANGE THIS
\usepackage{courier}  % DO NOT CHANGE THIS
\usepackage[hyphens]{url}  % DO NOT CHANGE THIS
\usepackage{graphicx} % DO NOT CHANGE THIS
\usepackage{natbib}  % DO NOT CHANGE THIS AND DO NOT ADD ANY OPTIONS TO IT
\usepackage{caption} % DO NOT CHANGE THIS AND DO NOT ADD ANY OPTIONS TO IT
\usepackage{algorithm}
\usepackage{algorithmic}

\usepackage{newfloat}
\usepackage{listings}
\DeclareCaptionStyle{ruled}{labelfont=normalfont,labelsep=colon,strut=off} % DO NOT CHANGE THIS
\floatstyle{ruled}
\newfloat{listing}{tb}{lst}{}
\floatname{listing}{Listing}
\gdef\showauthors@on{T}
\def\theauthors{\if T\showauthors@on\@author\else Anonymous submission\fi}

\usepackage{caption}
\usepackage{graphicx} %
\usepackage[hyphens]{url}  %
\usepackage{enumitem}

\usepackage{xspace}
\usepackage{multirow}
\usepackage{booktabs}
\usepackage{xcolor}
\usepackage[pagebackref,breaklinks,colorlinks=true,citecolor=blue]{hyperref}
\usepackage{amsfonts,amssymb,amsmath }
\newcommand{\argmax}{\mathop{\mathrm{argmax}}}

\usepackage{comment}
\usepackage[normalem]{ulem}
\usepackage{xcolor}
\usepackage{url}

\newcommand\XP[1]{\textcolor{black}{#1}}
\newcommand\XPC[1]{\textcolor{black}{[#1]}}

\newcommand\YB[1]{\textcolor{magenta}{#1}}

\title{Isolation and Impartial Aggregation: A Paradigm of Incremental Learning without Interference\footnote{This is the accepted version of \cite{wang2023isolation}, the Proceedings of the 37th AAAI Conference on Artificial Intelligence (AAAI23), Feburary 7–14, 2023, Washington DC, USA. \\ Please cite the final published version.}}

\author{Yabin Wang\textsuperscript{\rm1, 3}{\thanks{Yabin Wang and Zhiheng Ma are co-first authors.}}, Zhiheng Ma\textsuperscript{\rm 2 \dag}, Zhiwu Huang\textsuperscript{\rm 3}, Yaowei Wang\textsuperscript{\rm 4}, Zhou Su\textsuperscript{\rm 1}, Xiaopeng Hong\textsuperscript{\rm 5, 4, 1}{\thanks{Xiaopeng Hong is the corresponding author.}}\\
\textsuperscript{\rm 1}School of Cyber Science and Engineering, Xi'an Jiaotong University,
\textsuperscript{\rm 2}Shenzhen Institute of Advanced Technology, Chinese Academy of Science, \textsuperscript{\rm 3}Singapore Management University, Singapore, \textsuperscript{\rm 4}Peng Cheng Laboratory, \textsuperscript{\rm 5}Harbin Institute of Technology\\
{\tt\small iamwangyabin@stu.xjtu.edu.cn, zh.ma@siat.ac.cn, zzhiwu.huang@gmail.com}  \\
{\tt\small wangyw@pcl.ac.cn, \{zhousu, hongxiaopeng\}@ieee.org}}

\usepackage{bibentry}
\begin{document}

\maketitle

\begin{abstract}
%Incremental learning 

This paper focuses on the prevalent \XP{performance imbalance in the stages of} incremental learning. To avoid obvious stage learning bottlenecks\XP{, we propose a brand-new stage-isolation based incremental learning framework, which leverages a series of} stage-isolated classifiers to perform the learning task of each stage \XP{without the interference of others. To be concrete}, to aggregate multiple stage classifiers as a uniform one impartially, \XP{we firstly introduce a temperature-controlled energy metric for indicating the confidence score levels of the stage classifiers. We then} propose an anchor-based energy self-normalization strategy to ensure the stage classifiers work in the same energy level. \XP{Finally we} design a voting based inference augmentation strategy for robust inference.
%make these independent networks cooperate in prediction. 
%The alignment makes independent sub-networks have an equal energy level and thus can work together.
The proposed method is rehearsal free and can work for almost all continual learning scenarios. 
%\XP{such as class-incremental learning (CIL), domain-incremental learning, and cross-domain CIL.}
We evaluate the proposed method on four large benchmarks. Extensive results demonstrate the superiority of the proposed method in setting up new state-of-the-art overall performance.
\emph{Code is available at} \url{https://github.com/iamwangyabin/ESN}.

\end{abstract}

\section{Introduction}

\emph{Incremental learning} (\XP{\emph{a.k.a}}, continuous learning or lifelong learning) is a paradigm that continually evolves machine models on a data stream.
It is a longstanding research topic and might offer a path toward more human-like AI.
% without forgetting old knowledge they have learned before
% There are many areas in which deep neural networks have already advanced beyond human intelligence.
% However, human beings are capable of incremental learning a wide variety of tasks and performing impressively well, which is still challenging for deep networks.
% Thus, incremental learning (i.e., continuous learning or lifelong learning) has attracted much attention due to the ability to perform continuous model learning in a wide range of practical applications.
The stability-plasticity dilemma is {central to incremental learning~\cite{mai2022online,de2021continual}, which requires the models to be plastic to acquire new knowledge and stable to consolidate existing knowledge continuously.}

% The critical problem of incremental learning is \emph{catastrophic forgetting}\cite{mai2022online, hou2019learning}, a well-known phenomenon that knowledge learned before tends to forget after machine models learn new tasks.
% Previous works have to keep a fragile balance between stability and plasticity~\cite{mai2022online} to tackle this problem.
% That means machine models should be plastic to acquire new knowledge and stable to consolidate existing knowledge.
% However, the new tasks have indisputable advantages than previous tasks, and lead to server bias.
% The fragile balance is easily to break up and 
% However, most previous methods stay in a fragile balance between stability and plasticity.
% To tackle catastrophic forgetting, existing works on incremental learning usually use rehearsal memory or regularization terms to guide the update of machine models.

\begin{figure}[t] 
\centering
\includegraphics[width=0.4\textwidth]{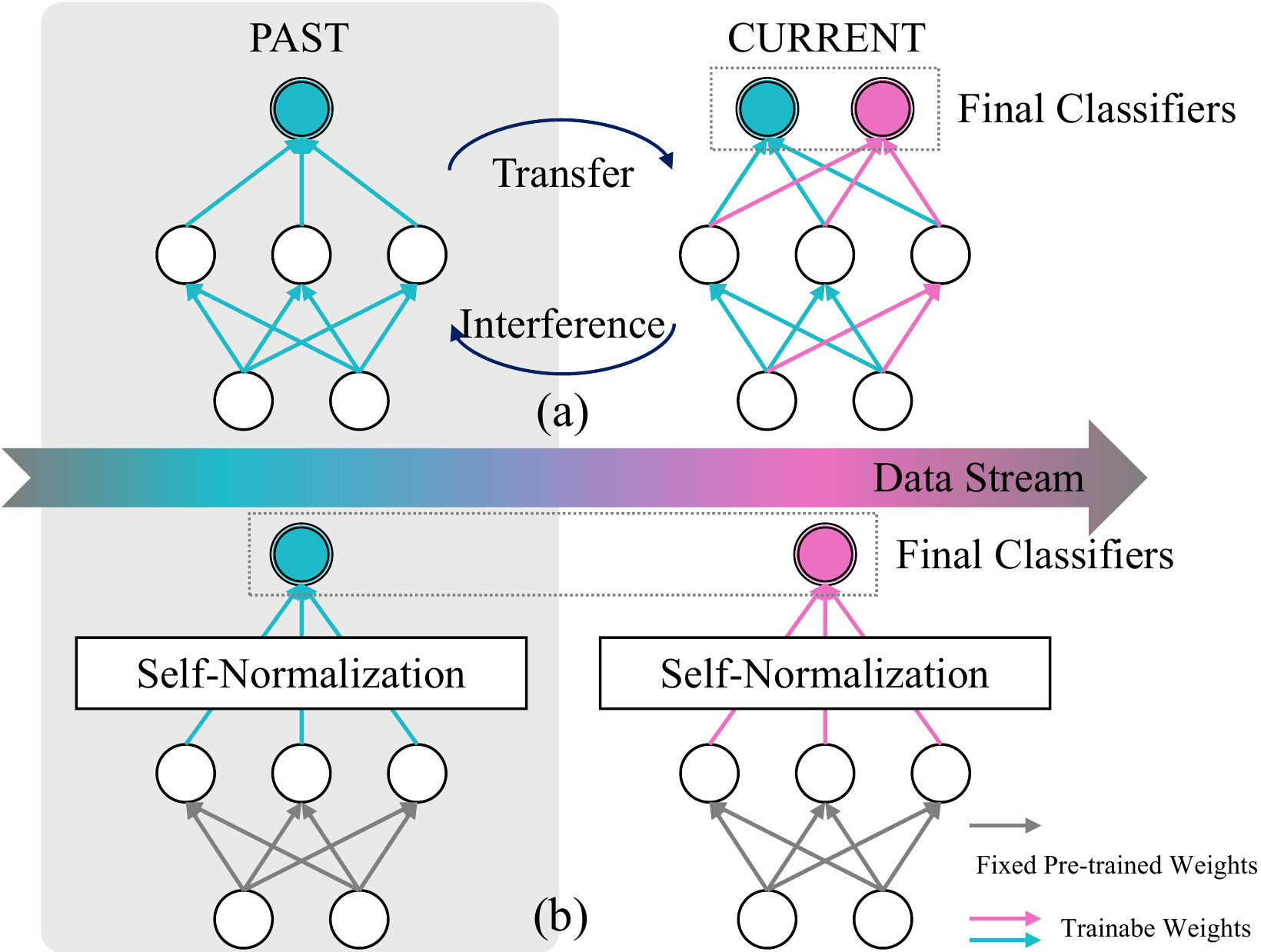} 
\caption{Comparison of the proposed ESN against traditional methods. (a) Existing methods usually use a uniform model to portray a heterogeneous data stream, \XP{which may cause performance imbalance and interference among the learning stages}. (b) In contrast, the proposed ESN uses a stage-isolation scheme for learning stage classifiers upon a fixed pre-trained backbone, resulting in \XP{much less forgetting and} interference.
} 
%, inevitably leading to more forgetting or interference
\label{fig:compare} 
\vspace{-0.5cm}
\end{figure}

Most previous works struggle to keep a fragile balance between stability and plasticity and also achieve pretty good results in terms of average accuracy, {which, however, results in tremendous performance gaps of different learning stages (a.k.a. sessions or tasks).}
%However, \XP{it can be easily found that the performance gaps} of different stages are tremendous. 
% For example, as shown in Fig.~\ref{fig:compare},  
%we measure the final accuracy of several widely compared methods. 
%Instead only report the final average accuracy, we use the stage-wise accuracies to study the hidden imbalance problem. 
% the stage-wise accuracy of BiC~\cite{wu2019large} and LwF~\cite{li2017learning} \XP{appear to have severe fluctuations, and the predictions are seriously biased towards newer classes. Similar phenomenon is reported in ~\cite{mai2022online}.}
This is a well-known phenomenon named class imbalance~\cite{mai2022online, de2021continual}. The source of this problem is twofold: firstly, the imbalance in the number of samples between the new  incoming data and the historical data;  secondly and more importantly, using a uniform model to portray a heterogeneous data stream \XP{may} result in \emph{a zero-sum game}~\cite{riemer2018learning,knoblauch2020optimal}, where one party gains mean another loses, as shown in Fig.~\ref{fig:compare} (a). This imbalance results in the breakdowns in recognizing certain classes, which creates a bottleneck in the final performance and limits the application of the model in real-world scenarios.
%This  problem~\cite{mai2022online} is due to the inherent contradiction of learning knowledge for a data stream and using a uniform classifier for prediction.
%Cram all knowledge into a single model without isolation generally result in \emph{a zero-sum game}~\cite{riemer2018learning,knoblauch2020optimal}, \XP{where one party gains mean another loses.}
%one side's gain is equivalent to the other's loss.
% We show the stage-wise final accuracy of three state-of-the-art methods in Fig.~\ref{fig:compare}.
% The final average accuracy of BiC~\cite{wu2019large} and LwF~\cite{li2017learning} is high, but their new stages' accuracies are much higher than previous ones.
%Just as the barrel theory shows, how much water the barrel can contain is not decided by the longest but the shortest board.
%This unfairness results in the accuracy collapsing of some categories, which makes the models incapable of being applied in real scenarios.
A few methods use a rehearsal buffer to alleviate such imbalance problems~\cite{hou2019learning}.
However, saving previous training data is memory expensive and has a privacy issue.
%Thus in this paper, we focus on the \emph{rehearsal-free general incremental learning} problem.

% This phenomenon, we call as  \emph{sequential unfairness}, is due to the inherent contradiction of keeping stability and plasticity at the same time.
% data imbalance of incremental learning, and thus hard to keep balance between new and old tasks. 

To address these issues, in this paper, we study how {to create \emph{win-win} solutions for all-stage learning}.
%based on aggregation.
{We challenge the traditional unified paradigm and suggest a  stage-isolation scheme for learning stage classifiers (in Fig.~\ref{fig:compare} (b)).} \emph{Stage isolation} is targeted at learning multiple high confidence and low bias {stage-specific} classifiers for every stage \XP{in isolation}, so that the classifier in each stage can be shielded from the interference of other stages, to satisfy the performance requirements of each stage adequately.

Nevertheless, the main difficulty of this paradigm lies in how to aggregate in an \emph{impartial} way multiple isolated learners trained on different stages, as the learners trained on different stages of various incoming data stream \XP{separately} may have diverse class-wise confidence distributions. For example, as shown by Fig. 3(b), the output scores of the classifiers of two stages can be clearly different. A straightforward aggregation like finding the class with the highest confidence still has a tendency to \emph{stage imbalance}. %Through the analysis provided, we realize that the main difficulty of this paradigm} lies in how to aggregate in an \emph{impartial} way multiple independent learners trained on different stages. 

We contend that the key to solving this problem lies in the regularization of the stage classifier outputs. Specifically, there are three criterions to meet. \textbf{Criterion 1}: the stage classifier should have higher output confidence scores for the data within the stage it belongs to (i.e., in-stage data) than others (i.e., out-stage data);
\textbf{Criterion 2}: the confidence scores for in-stage data should be consistent across all stages; \textbf{Criterion 3}: the right stage classifier for the in-stage data shall have the highest confidence score among all classifiers. Unfortunately, they are challenging to be satisfied in \XP{the incremental learning scenarios}. The reasons are two-fold. {First, optimizing the stage learners only using the current data (the only one accessible) will result in a serious bias. Second, it is impossible to make full regularization as the classifiers to be learnt in future can not be considered at current stage.}
%as stage classifiers are incrementally added, we cannot consider future classifiers when performing regularization.}

Dealing with such a \emph{backward-compatible regularization} dilemma,  inspired by the Helmholtz free energy theory~\cite{lecun2006tutorial}, we introduce a temperature controlled energy metric to reflect the confidence score levels of stage classifiers. On this basis, we provide a \emph{rehearsal-free}  incremental learning paradigm, which regularize the stage classifiers for aggregating them impartially as a uniform classifier.  
Specifically, we first use the \XP{pre-trained ViT~\cite{dosovitskiy2020vit}  backbone as a frozen strong prior before the stage-specific classifiers to assure higher confidence scores for} the in-stage data than the out-stage data as far as possible, which has been proved in~\cite{fort2021exploring} (for \textbf{Criterion 1}). Second, we design an anchor-based energy self-normalization loss, \XP{which} restricts the energy metrics of stage classifiers tightly around the energy anchor, to ensure all stage classifiers lay in the same energy level {when facing in-stage data of their own (for \textbf{Criterion 2}). %Combining \textbf{Criterion 1} and \textbf{Criterion 2}, we can roughly guarantee \textbf{Criterion 3}.
Furthermore, though the `so far' best control parameter for the current stage can be found by a design method, it works only in a backward-compatible manner. 
 To avoid overfitting to any specific stage, we maintain the `so far' best parameters for all stages met and use a voting scheme to produce reliable inference outputs, by which \textbf{Criterion 3} can be better approached by stages\footnote{\textbf{Criteria 1} and \textbf{2} together form a rough guarantee of \textbf{Criterion 3}, as shown in the `Proposed Method' Section.}.}
 
 %We then regularize the stage classifiers to restrict their energy metrics tightly around the energy anchors, by which \textbf{Criterion 2} can be met by stages. 
 %Finally, as the control parameter is found only in a backward compatible manner, to avoid overfitting to any specific stage, we maintain the `so far' best parameters for all stages met and use a voting scheme to produce reliable inference outputs.}
 
%The proposed rehearsal-free incremental method \XP{can handle} almost all scenarios, including class-incremental learning (CIL), domain-incremental learning (DIL), and cross-domain class incremental learning. 
%\XP{To further facilitate the research on this problem,} we further propose a challenging benchmark for cross-domain class incremental learning.
%Experiments on four large benchmarks show that the performance of the proposed method \XP{sets up new} state-of-the-art results.

To summarize, we propose a brand-new rehearsal-free general incremental learning paradigm to tackle the \XP{performance} imbalance and the zero-sum game
problems, called \textbf{E}nergy \textbf{S}elf-\textbf{N}ormalization (\textbf{ESN}), which {can handle} almost all scenarios, including class-incremental learning (CIL)~\cite{de2021continual}, domain-incremental learning (DIL)~\cite{wang2022s}, and cross-domain class incremental learning~\cite{xie2022general}. The contributions can be further detailed as follows:

\begin{itemize}[noitemsep,topsep=0pt,parsep=2pt,partopsep=0pt,leftmargin=1em]%\XP

\item We propose the {anchor-based energy self-normalization (ESN)} {so that stage-classifiers can produce high and consistent confidence scores for in-stage data.}
%aggregate multiple independent classifiers.
\item {We design a control parameter (temperature) finding method to obtain stage-cumulative best parameters for progressively ensuring `right' classifiers with the highest scores. On this basis, we propose a voting based inference augmentation strategy for robust inference.}
%\item \XP{We propose a voting based inference augmentation strategy for robust inference.}
%\item \XP{To facilitate the research on this problem,} we further propose a challenging benchmark for cross-domain class incremental learning.
\item \XP{The proposed ESN sets up new state-of-the-art performance, as shown by extensive experiments on four large-scale benchmarks. A challenging benchmark for cross-domain class incremental learning is built as well.}

%\item Our method gets state-of-the-art results on four large public benchmarks.
\end{itemize}
\section{Related works}

\subsection{Incremental Learning}

\begin{comment}
%Incremental learning has been studied for years.
%The main challenge is the \emph{catastrophic forgetting} phenomenon, which means the knowledge learned before tends to forget after learning on new tasks~\cite{de2021continual}.
There are three main  scenarios.
Task incremental learning learns new tasks incrementally, with the task identification  provided during inference~\cite{pourkeshavarzi2021looking}.
%, which usually regarded as a simple scenario. 
Domain incremental learning~\cite{wang2022s} learns new tasks from different domains, and Class incremental learning~\cite{de2021continual} learns new classes in sequence.
In this paper, we aim to provide a general solution to incremental learning.
%tackle these three common scenarios at the same time.
\end{comment}

There are three main types of incremental learning methods~\cite{de2021continual}.

\emph{Rehearsal-based methods} tackle catastrophic forgetting either by keep a small set of old training examples in memory~\cite{tao2020topology, dong2021few, liu2022model} or using synthesized data produced by generative models~\cite{shin2017continual}.
By using rehearsal buffer for knowledge distillation and regularization, rehearsal-based methods have achieved state-of-the-art results on various benchmarks~\cite{douillard2022dytox,joseph2022energy, 9932643}.
% To make full use of saved exemplars, some works propose complicated regularization strategies~\cite{, dong2021few, } and more balance classifiers~\cite{hou2019learning}.
However, the performance of rehearsal-based methods generally deteriorates with smaller buffer size~\cite{mai2022online}. 
% and saving small amount of old data also leads to more server data imbalance problems~\cite{ hou2019learning}.
What's more, it is often more desired that the exemplars of old tasks are not stored for the data security and privacy \XP{reasons}~\cite{wang2022s}. 

\emph{Regularization-based methods} design knowledge distillation strategies~\cite{li2017learning} or parameter regularization terms~\cite{kirkpatrick2017overcoming} to mitigate catastrophic forgetting.
% Due to the lack of solid reference data (i.e, exemplars), it is hard to a fragile balance between stability and plasticity.

\emph{Network-based methods} modify networks' architecture during incremental learning to mitigate catastrophic forgetting.
Some works expand network parameters to learn new tasks and get state-of-the-art performances~\cite{yan2021dynamically, Wang_2022_CVPR, xu2020aanet, douillard2022dytox}.
Also some methods use the parameter isolation strategy to keep each task independent~\cite{serra2018overcoming, li2019learn}.
Recently, L2P~\cite{Wang_2022_CVPR}  uses prompt tuning and pre-trained models for incremental learning tasks.  
Parameter efficient fine-tuning, like prompt tuning, offers a promising way for incremental learning problems.
However, L2P is still a unified-structure model, and it requires a fixed query function to find prompts, which is \XP{time-consuming and less efficient in} complicated situations.
% The main problems are that the parameter number of new network cannot be too large, and how to combine multiple independent networks for prediction.

\subsection{Energy-based Models}

Energy-based Models (EBMs)~\cite{lecun2006tutorial} capture dependencies of variables by associating a scalar energy to each configuration of the variables.
EBMs have been used for generative modeling~\cite{du2019implicit}, out-of-distribution detection~\cite{liu2020energy}, and open-set classification~\cite{al2022energy}.
Despite being successful across various tasks, EBMs have limited applications in incremental learning. %and all relay on rehearsal buffer to work. 
ELI~\cite{joseph2022energy} proposes to learn an energy manifold to counter the representational shift that happens during incremental learning.
% The energy manifold proposed by ELI can augment the stored samples for knowledge distillation.
It uses EBMs to portray changes of the model and then try to compensate the updated model to the original one, which is still a tug of war.
What's more, it assumes the energy between different stages is distinguishable, which is a \XP{\emph{too} strong assumption in application scenarios}.
EA~\cite{zhao2022energy} also uses energy-based model to add the calculated shift scalars onto the output logits to mitigate class imbalance.
The calculation of compensation scalars is based on the samples of all classes, which \XP{suggests that it relies} on rehearsal buffer.
% EA directly adds class-wise scalars to the model's output to let the prediction lean to the other side, which must need the saved old data to compute compensate scalars for each class.
Both works still struggle to alleviate the imbalance problem in a uniform model.
Moreover, they are both rehearsal-based and can only handle CIL problems, which are \XP{far from} general and robust solutions for incremental learning.

\section{Proposed Method}

\subsection{Problem Definition}

Incremental learning refers to training the model in a data stream, while the model can only access part of the training data at a time.
Let $\zeta=\{1,2,3, ..., S\}$ denote the Stage-ID set, where $S$ is the current maximum stage number. The incoming data of the $s$-th stage is denoted as $\mathcal{D}^s = \left \{  x_i,y_i \right \}_{i=1}^{N^s}$, where  ${N^s}$ is the total samples number of this stage. $\left (x,y \right ) \sim p_{data}^{s}$ represents the data distribution of the $s$-th stage.
For class incremental learning, different stages have different categories to learn, and there is no category overlap, $\mathcal{Y}^{i} \cap \mathcal{Y}^{j}=\emptyset$, where $\mathcal{Y}^{s}$ is the label set of the $s$-th stage.
For domain incremental learning, the categories maintains the same for all stages, $\mathcal{Y}^{i} = \mathcal{Y}^{j}$, but data distribution of each stage is different or even highly heterogeneous.

Our proposed ESN can handle these two challenging scenarios at the same time and even more challenging cross-domain class incremental learning, in which different stages have different categories from different domains.

\subsection{Overall Framework}

% The out-of-distribution (ood) task is to determine whether the test data are drawn from the same distribution $p_{in}$ as the training data, which can be consider as a binary classification problem:
% \begin{equation}
% G(x)=
% \begin{cases}
% \text{In}& H(x) >= \tau \\
% \text{Out}& H(x) < \tau 
% \end{cases},
% \end{equation}
% where $G(x)$ is a binary classifier, $H(x)$ is the confidence score function mapping inputs to scalars, and $\tau$ is the pre-defined threshold. 

% As can be seen, our methods and confidence score functions, which are trained \textbf{independently} with the corresponding data $\mathcal{D}^s$, without accessing other sessions' data. Since the confidence score is distinguishable between in-distribution and out-of-distribution data, theoretically we can determine the Session-ID of the input by comparing confidence scores between different classifiers,

% For example, if the lowest confidence score of one session's out-of-distribution data is higher than the highest score of another session's in-distribution data, then the Session-ID prediction will always be the first one, even the out-of-distribution detection accuracy of each session is very high.

Previous incremental learning methods need to find a fragile balance between stability and plasticity. Using a uniform model to portray a heterogeneous data stream may result in a zero-sum game and be seriously biased toward newer classes~\cite{de2021continual, mai2022online}.

% However, due to the data imbalance (rehearsal-based methods~\cite{wu2019large,buzzega2020dark,cha2021co2l}) or the continual weight updating problem in a unify model (regularization-based methods~\cite{li2017learning,kirkpatrick2017overcoming}), the new tasks still have indisputable advantages than previous tasks, we name this phenomenon as \textbf{sequential unfairness}. 
% For scenarios where the former and later tasks have equal importance, the sequential unfairness is unacceptable. 
% We find out that although the average accuracy of previous methods seems acceptable, the sequential unfairness results in much lower accuracy of earlier tasks (as shown in Fig.~\ref{fig:compare}), which makes the models incapable to be applied in real scenarios. 
% Just as the barrel theory shows, how much water the barrel can contain is not decided by the longest but the shortest board.

\begin{figure}[] 
\centering
\includegraphics[width=0.45\textwidth]{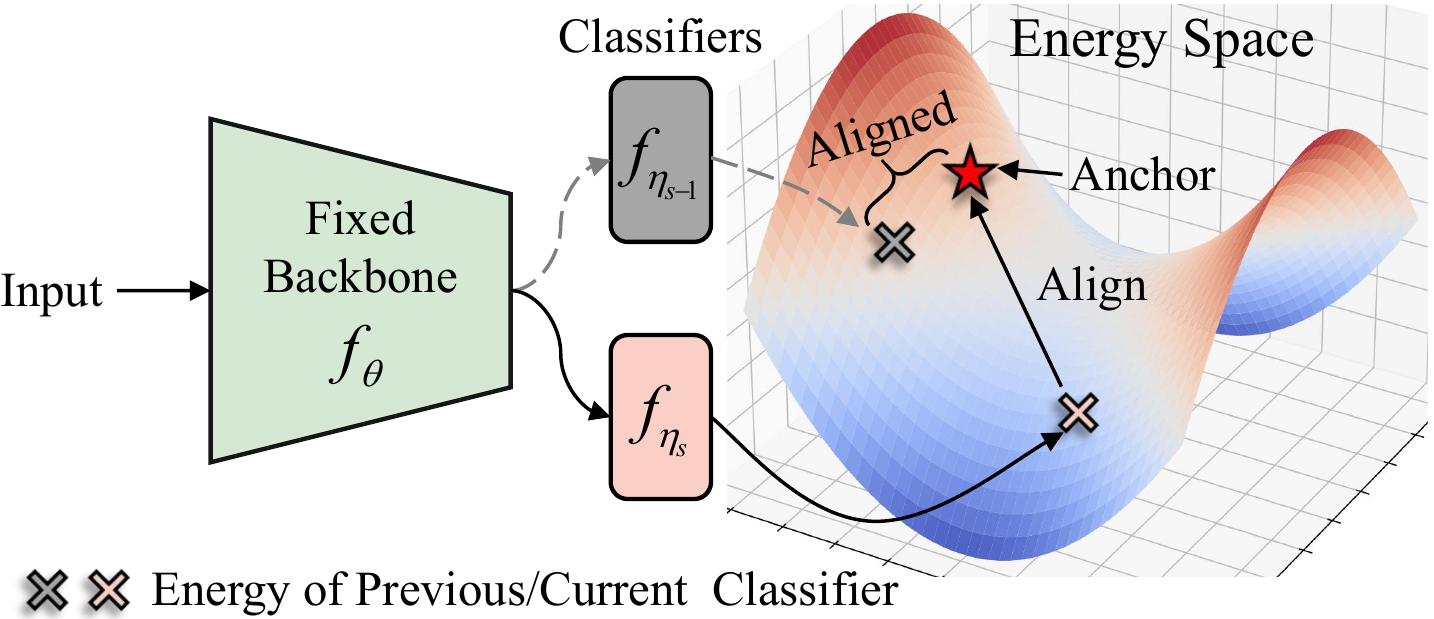} 
\caption{\XP{Overview of the proposed anchor-based energy self-normalization for stage classifiers. The classifiers of the current and the previous stages, $f_{\eta_s}$ and $f_{\eta_{s-1}}$, are aligned sequentially to restrict their energies around the anchor.}
%Overview of the proposed stage classifier energy self-normalization. ESN uses an anchor-based energy self-normalization to restrict energies of all isolated classifiers $f_{\eta_s}$ lay around the anchor stage-by-stage.
% So that we can aggregate all classifiers trained independently to get the final prediction.
% We train independent headers for every stage, which share the same backbone network. For a given input sample $x$, each header will yield logits.
% At the same time, we aggregate their confidence score $H^s(x)$ to predict the most confidence stage identification $s^*$.
% The final prediction $y^*$ is the argmax output of the corresponding header.
} 
\label{fig:overview} 
\vspace{-0.5cm}
\end{figure}

In this paper, we proposed a brand-new rehearsal-free general incremental learning paradigm to tackle the imbalance and the zero-sum game problems. 
% The framework of our proposed method is as shown in Fig.~\ref{fig:overview}.
Specifically, we train multiple isolated stage-specific classifiers upon a frozen pre-trained backbone for each stage. 
% In the training stage, each stage's classifier is trained independently on the corresponding data without noticing other stages' data and classifiers. 
In the inference phase, we first select the most confident classifier (Eq.~\ref{eq:overview_step1}), and then use it to predict the final result (Eq.~\ref{eq:overview_step2}).

%  as Eq.~\ref{eq:overview}.

% \begin{equation}
% \begin{aligned}
% \text{STEP 1: \quad}& s^* = \argmax_{s \in \zeta}H^{s}(x), \\
% \text{STEP 2: \quad}& y^* =  \argmax_{y \in \mathcal{Y}^{s^*}}P^{s^{*}}(y|x),
% \end{aligned}
% \label{eq:overview}
% \end{equation}

\begin{equation}
   s^* = \argmax_{s \in \zeta}H^{s}(x),
\label{eq:overview_step1}
\end{equation}

\begin{equation}
   y^* =  \argmax_{y \in \mathcal{Y}^{s^*}}P^{s^{*}}(y|x),
\label{eq:overview_step2}
\end{equation}
where $P^{s}(y|x)$ is the classifier of the $s$-th stage, and its confidence score function is denoted as $H^{s}(x)$. As shown, comparability between different stages' confidence scores is the guarantee of impartial aggregation.

% The confidence score function is widely used in problems such as out-of-distribution detection~\cite{wang2022vim}, uncertainty estimation~\cite{mehrtash2020confidence}, novelty detection~\cite{sun2022gradient}. 
% It assumes that the confidence score should be high when the test data is from the same distribution as its training set. Otherwise, when the test data is from other distributions, the confidence score should be low. 

% Unlike previous methods that use a single uniform model, our methods consist of multiple classifiers trained independently.
As shown in Fig.~\ref{fig:compare}, given a pre-trained backbone $f_{\theta}$, each stage we initialize a specific classifier $f_{\eta_s}$.
During training at stage $s$, we freeze the pre-trained backbone $f_{\theta}$ and only update parameters of the classifier $\eta_s$.
$\theta$ and $\eta$ are parameters of backbone and classifier respectively.
For simplicity, we use the ViT-B/16 pre-trained on ImageNet as freeze backbone, and use the class-attention block (CAB)~\cite{touvron2021going} with a linear projection as the classifier.
Our proposed strategy is also suitable for other parameter isolation methods~\cite{jia2022vpt}, which we will show later in the experiments.
The stage isolated classifier can shield the interference of each other.

% Our proposed methods can be applied to many network-based incremental learning methods, and the key idea is to learn independent classifiers for each stage.

% In the following sections, we first introduce the training method based on energy alignment to ensure fairness and comparability between different classifiers, and then introduce the confidence score function and our aggregation inference strategy in detail.

In the following sections, we first detail the training method based on the self-normalization strategy, which ensures the impartial aggregation of all stages' classifiers. Then we introduce the stage-cumulative control parameter optimization method with voting-based inference augmentation to further promote the performance.

\subsection{\XP{Stage Classifier Self-Normalization}}

The most commonly used training criterion in training deep neural networks is the softmax cross-entropy loss. 
However, previous works~\cite{tang2021codes,liu2020energy} show that directly training with this loss results in overconfidence issues, where the maximum softmax activation value always approaches one in despite of the data is from training data distribution or not. Previous works have shown that other criteria such as the Helmholtz free energy~\cite{liu2020energy} or the maximum logit value~\cite{hendrycks2019scaling} are better confidence scores than the maximum softmax value. However, \textbf{none of the above works discuss how to align confidence scores between different classifies learned from data steam}. 

Next, we first briefly review the relationship between the softmax cross-entropy loss and the energy-based model~\cite{grathwohl2019your,liu2020energy,lecun2006tutorial}, then propose the anchor-based energy self-normalization objective function, which makes energy for in-stage data consistent across stages.

% then propose the energy alignment regularization term based on a shared energy anchor between different classifiers. 
% The alignment loss can make classifiers trained with different data distribution lay in the same energy plane.

Let's define the energy function for a given input-label pair $(x,y)$ as follows:

\begin{equation}
E^{s}(x, y) = -h^{s}(x)[y],
\label{eq:energy_x}
\end{equation}
where $h^s(x) = f_{\eta_s}(f_{\theta}(x))$ is the logits of the $s$-th classifier, and $h^s(x)[y]$ is the logit value of $y \in \mathcal{Y}^{s}$, then softmax activation can be considered as a special case of discrete Gibbs distribution when the temperature parameter $T$ equals to $1$:

\begin{equation}
P^{s}_{T}(y|x) = \frac{\exp(-E^{s}(x, y)/T)}{\exp(-\mathcal{F}^{s}_{T}(x)/T)},
\label{eq:probability density}
\end{equation}
where $\mathcal{F}^{s}_{T}(x)$ is the Helmholtz free energy, which can be expressed as the negative log partition function:

\begin{equation}
\mathcal{F}^{s}_{T}(x) = -T\log{\sum_{y \in \mathcal{Y}^{s}} \exp{(-E^{s}(x, y)/T)}}.
\label{eq:freeenergy}
\end{equation}

% When only part of data is visible, the classifier trained with the negative log likelihood will increase the confidence of the currently visible categories, while suppress the confidence of the previous category.
% On the contrary, keeping the superiority of old categories will suppress new knowledge.
% This is the sequential unfairness.
% In order to mitigate this imbalance problem, previous work also designs various regularization terms or classifiers (i.e, NME).
% For our independent classifiers, the imbalance problem will be more severe.
% Because each classifier can only access a part of the data, different classifiers are significantly different, and it is almost impossible to directly combine them to get the maximum value.

% Here we detail our elegant and effective energy alignment method to align all independent classifiers and make prediction.
% The training process is completely independent and is almost the same as general classification tasks.
% Let . We use Cross-entropy to optimize session independent classification models, which is commonly used in machine learning.

Thus, the softmax cross-entropy loss can be rewritten as Eq.~\ref{eq:celoss}.

\begin{subequations}
\label{eq:celoss}
\begin{align*}
   \mathcal{L}_{ce}^{s} &= \mathbb{E}_{(x,y) \sim p^{s}_{data}}(-\log{P^{s}_{T}(y|x)}) \nonumber\\
    &= \frac{1}{T}\mathbb{E}_{(x,y) \sim p^{s}_{data}}(E^{s}(y,x)-\mathcal{F}^{s}_{T}(x)). \tag{6}
\end{align*}
\end{subequations}

\begin{figure}[t] 
\centering
\includegraphics[width=0.5\textwidth]{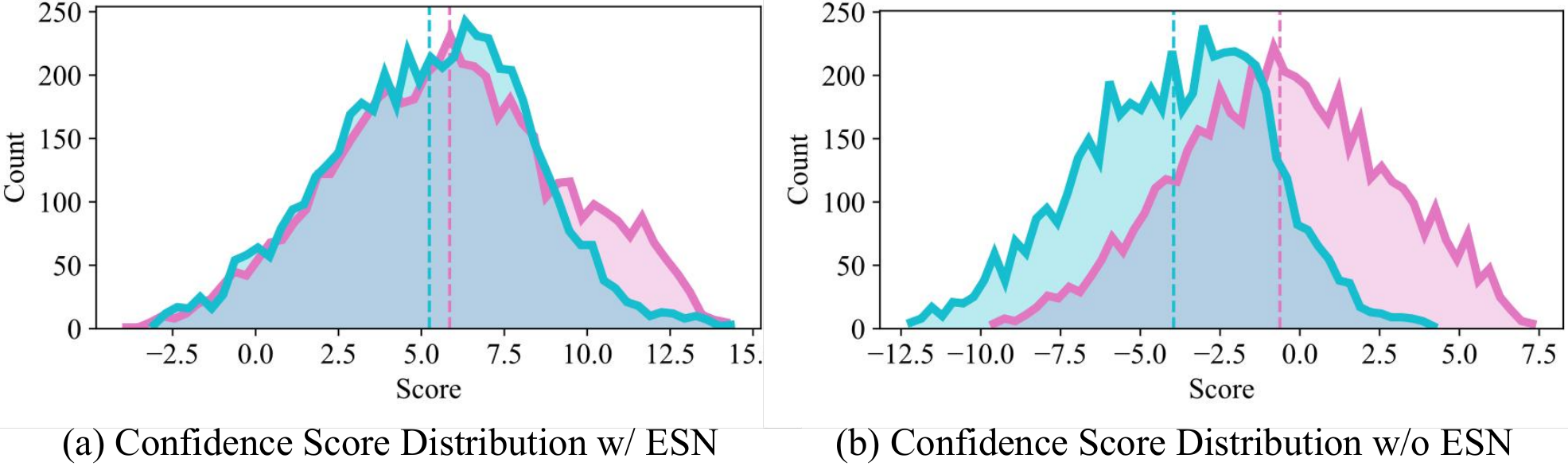} 
\caption{Distribution shift. We extract $2$ stages' training data from Split CIFAR-100 and show their confidence scores trained with and without using our proposed anchor-based energy self-normalization. The y-axis is the count of image, and the x-axis is the confidence score.} 
\label{fig:distcomp} 
\vspace{-0.5cm}
\end{figure}

As can be seen, the softmax cross-entropy loss will decrease the energy between the input data and the ground-truth label while increasing the overall Helmholtz free energy. 
However, when $E^{s}(y,x)$ and $\mathcal{F}^{s}_{T}(x)$ are added with the same scalar, the loss value remains unchanged, which makes it meaningless to directly compare the free energy between different classifiers trained independently using softmax cross-entropy loss. 
To fix this issue, we propose a simple but effective energy self-normalization loss $\mathcal{L}_{al}^{s}$, which constrains the free energy of each classifier with a fixed anchor $\Delta$, as Eq.~\ref{eq:anchorloss}.

\begin{equation}
   \mathcal{L}_{al}^{s} = \mathbb{E}_{x \sim  p^{s}_{data}}(\mathcal{F}^{s}_{T}(x)-\Delta)^{2},
\label{eq:anchorloss}
\end{equation}
where $\Delta$ is a preset hyper-parameter, and the experimental results show that ESN is insensitive to its value. 
The total loss trained for every individual classifiers is given by Eq.~\ref{eq:totalloss}.

\begin{equation}
   \mathcal{L}_{total}^{s} =  \mathbb{E}_{(x^{s},y^{s}) \sim  p^{s}_{data}}(\mathcal{L}_{ce}^{s}+\lambda\mathcal{L}_{al}^{s}),
\label{eq:totalloss}
\end{equation}
where $\lambda$ is a hyper-parameter to balance $\mathcal{L}_{al}^{s}$ term. 
And we choose a representative temperature $T=1$ during training.
% It's worth to mention that since our classifier is trainable, the temperature parameter $T$ in the training phase is not important, so we directly set it to $1$. 
Our complete training algorithm is introduced in Alg.~\ref{alg:algo_train}. Fig.~\ref{fig:distcomp} visualizes the free energy distribution with and without the self-normalization, which illustrates the effectiveness of ESN.

\begin{algorithm}[thbp!]
\begin{algorithmic}[1]
\item \textbf{Given components:} Pre-trained backbone $f_{\theta}$, stage classifier $f_{\eta}$, total stage number $S$, training iterations for each stage $M$, energy anchor $\Delta$, training data  $\mathcal{D}$, learning rate $\epsilon$, temperature pool $\Omega$, candidate temperature pool $\Psi$;
\FOR{ $s = 1,\cdots,S$} 
  \STATE Initialize classifier $f_{\eta_s}$ for the stage $s$;
  \FOR {$m = 1,\cdots,M$}
    \STATE Draw a mini-batch training data $B$ from $\mathcal{D}_s$;
    \STATE Calculate logits $h^s(x)= f_{\eta_s} ( f_{\theta} (x))$;
    \STATE Calculate loss $\mathcal{L}_{total}^{s}$ by Eq.\ref{eq:totalloss};
    % \item Update $\tau_s$ by $\tau_s \gets \tau_s - \epsilon \bigtriangledown_{\tau_s} \mathcal{L}_B $
    \STATE Update $\eta_s$ by $\eta_s \gets \eta_s - \epsilon \bigtriangledown_{\eta_s} \mathcal{L}_{total}^{s} $;
  \ENDFOR
  \IF{$s>1$} 
  \FOR{$t \in \Psi$}
  \STATE Extract Helmholtz free energy $-\mathcal{F}^{s}_{T}(x)$ by Eq.~\ref{eq:scaling};
  \STATE Calculate stage identification by $s^* = \argmax_{s \in \zeta}(-\mathcal{F}^{s}_{T}(x))$;
  \STATE Calculate stage identification accuracy $ACC_t$ of temperature $t$ by $\sum (s^* == s)$;
  \ENDFOR
  \STATE $\Omega \gets \argmax_{t \in \Psi} ACC_t$;
  \ENDIF
\STATE Return the model parameters $\eta_s$.
\ENDFOR
\end{algorithmic}
\caption{{Model Training}}
\label{alg:algo_train}
\end{algorithm}

\subsection{Voting with Stage-Cumulative Temperatures}

As we have already normalized the Helmholtz free energy with the fixed anchor (Eq.~\ref{eq:anchorloss}), taking the negative Helmholtz free energy as the confidence score is a natural choice:
\begin{equation}
H^{s}(x) = -\mathcal{F}^{s}_{T}(x) = T\log{\sum_{y \in \mathcal{Y}^{s}} \exp{(h^s(x)[y]/T)}},
\label{eq:scaling}
\end{equation}
which is the \textit{logsumexp} of the logits with the control temperature parameter $T$. Previous energy-based out-of-distribution detection methods~\cite{wang2022vim, liu2020energy} have shown that the in-distribution data usually has a lower free energy (higher confidence scores) than the out-of-distribution data for a certain classifier. Further aided by the energy self-normalization objective function, we can approximately derive that the right stage classifier for the in-stage data shall have the highest confidence score among all classifiers (Eq.~\ref{eq:overview_step1}). The derivation can be briefly expressed as $H^{i}(x^{i}) = H^{j}(x^{j}),H^{j}(x^{j}) > H^{j}(x^{i}) \rightarrow H^{i}(x^{i}) > H^{j}(x^{i})$, where $x^{i}$ is the in-stage data of the $i$-th stage but the out-stage data of the $j$-th stage. However, this derivation only approximately hold, we further propose a stage-cumulative temperature calibration strategy with a voting inference augmentation to further optimize the maximum confidence criterion (Eq.~\ref{eq:overview_step1}) without overfitting the newest stage's data.

% During training, we set the temperature $T$ to be $1$, and align classifiers to the same energy plane within in-stage data.
% However, only in-stage data will inevitably make classifiers overfit to their domain.
% Based on the Helmholtz free energy, we further propose a post-hoc optimization strategy to improve the accuracy with negligible computation overhead. 

As shown in Eq.~\eqref{eq:freeenergy}, we can adjust the free energy by changing the temperature parameter $T$. Theoretically, we can find out the optimal temperature $T$ for each classifier by optimizing the stage-ID prediction accuracy with all stages' data, which is not possible in rehearsal-free incremental learning. As we can only access the current stage's data, we propose a stage-cumulative strategy to avoid overfitting.

Firstly, we find out the optimal temperature only with the current stage's training data by traversing the candidate temperatures $\Psi$ and choosing the one with the best stage-ID prediction accuracy of the current stage.
Secondly, we add this temperature to the final temperature pool denoted as $\Omega$.
Finally, we can traverse temperatures in the temperature pool $\Omega$, and then aggregate stage-ID predictions under different temperatures by voting.
To keep the fairness and the comparability between different classifiers, we simultaneously change the temperature for all classifiers, and do voting as Eq.~\ref{eq:votinge}.
\begin{equation}
s^* = \text{MODE}(\{\argmax_{s \in \zeta} -\mathcal{F}^{s}_{T}(x) |\text{For} \ \  T \in \Omega \}),
\label{eq:votinge}
\end{equation}
where $\text{MODE}(\cdot)$ is the mode operator to find the most frequent element in a collection. 
This voting-based inference augmentation strategy only increases negligible computation overhead. After the logits predicted by the model by once, we only need to recalculate Eq.~\eqref{eq:scaling} under different $T$.

Our augmented inference algorithm is introduced in Alg.~\ref{alg:algo_test}, and the stage-cumulative temparature calibration is already introduced in Alg.~\ref{alg:algo_train}.

\begin{algorithm}[t]
\begin{algorithmic}[1]
\item \textbf{Given components:} Pre-trained backbone $f_{\theta}$, stage classifiers $\{f_{\eta_s}\}_{s=1}^S$, temperature pool $\Omega$, total stage number $S$; 
\STATE \textbf{Input:} Test example $x$; 
\STATE Calculate image feature $l(x) = f_{\theta}(x)$; 
\FOR{$s = 1,\cdots,S$} 
  \STATE Generate $s$-th logits $h^s(x)= f_{\eta_s} (l(x))$;
  \FOR {$t \in \Omega $}
    \STATE Calculate scaled energy $-\mathcal{F}^{s}_{T}(x)$ by Eq.~\ref{eq:scaling};
  \ENDFOR
\ENDFOR
\STATE Voting for stage identification $s^*$ by Eq.\ref{eq:votinge};
\STATE Return final prediction $y^*$ by Eq.~\ref{eq:overview_step2}.
\end{algorithmic}
\caption{\XP{Inference}}
\label{alg:algo_test}
\end{algorithm}

\section{Experiments}

\subsection{Benchmarks and implementation}

% \XP{To explain why we need to build benchmark.}

We conduct extensive experiments to evaluate the proposed ESN.
We consider two main incremental learning scenarios: (1) class-incremental learning where classes are generally from the same domain ; (2) domain-incremental learning where classes are the same but from different domains.
Moreover, we consider a more general scenario: the cross-domain class incremental learning, where different classes from diverse domains.
And build a benchmark, named as Split DomainNet, for this scenario.

We evaluate ESN on CIFAR-100~\cite{krizhevsky2009learning}, Split DomainNet, 5-datasets~\cite{ebrahimi2020adversarial} and CORe50~\cite{lomonaco2017core50}.

\noindent
\textbf{Split DomainNet Benchmark}
We build the cross-domain incremental learning benchmark, Split DomainNet, based on DomainNet~\cite{peng2019moment}.
The Split DomainNet is the scenerio that incoming data of each stage contains images of new categories from different domains.
We construct this dataset as a benchmark for the cross-domain class incremental learning, which we believe is a more challenging and practical scenario.
DomainNet collects images of $345$ common objects from $6$ diverse domains including Clipart, Real, Sketch, Infograph, Painting and Quickdraw.
% This is a popular and challenging benchmark for domain generalization even using well pre-trained models.
% We propose to use this dataset as a benchmark for a more general incremental leaning scenario.
Because some domains and categories in DomainNet contain few instances (even without a single instance), we select the top $200$ categories with the most images.
We split the $200$ classes randomly into ten stages with $20$ classes per stage.
Instances of each stage come from a randomly selected domain.
% There are also some incremental learning methods use DomainNet for evaluation~\cite{xie2022general, simon2022generalizing}.
% However, these settings mainly evaluate the model's domain generalization during incremental learning, and the exact setting is not publicly available.
% Additionally, there is a lack of a large evaluation benchmark for incremental learning using pre-trained models, similar to the commonly used ImageNet benchmark.
% What's more, for incremental learning methods based on pre-trained models, commonly used ImageNet benchmark is improper.
% Thus we release Split DomainNet as a more challenging and practical benchmark for incremental learning.

\noindent
\textbf{Split-CIFAR100 Benchmark} CIFAR-100~\cite{krizhevsky2009learning} is a widely used benchmark for class-incremental learning.
% , which has $60,000$ RGB images over $100$ classes. And the image size is $32 \times 32$.
Split CIFAR-100 splits the origin CIFAR-100 into $10$ sessions and each session has $10$ classes. 

\noindent
\textbf{5-Datasets Benchmark}
5-Datasets~\cite{ebrahimi2020adversarial} is a benchmark for class incremental learning.
% The sequence of 5-Datasets includes SVHN~\cite{netzer2011reading}, CIFAR10~\cite{krizhevsky2009learning}, not-MNIST, Fashion-MNIST~\cite{xiao2017fashion} and MNIST~\cite{lecun1998gradient}.
Although each dataset in 5-Datasets is not difficult, it is still a challenging benchmark for pre-trained models, because there are slight similarity between them.

\noindent
\textbf{CORe50 Benchmark}
CORe50~\cite{lomonaco2017core50} is a large benchmark for continual object recognition.
This dataset collects images of $50$ different objects from $11$ distinct domains ($8$ indoor and $3$ outdoor).
Three domains (3, 7, and 10) are selected as test set, and the remaining $8$ domains are used for incremental learning.
CORe50 is a benchmark for domain-incremental learning.

\noindent
\textbf{Evaluation Metrics.}
We use the Final Average Accuracy (FAA) and Final Forgetting (FF) as evaluation metrics for class-incremental learning and cross-domain task incremental learning, which are widely used in previous works~\cite{mai2022online}.
There is no distinct task boundary for Domain-incremental learning, and we use the Final Average Accuracy (FAA).

\begin{table}[] \small
\setlength{\tabcolsep}{10pt}
\begin{center}
\begin{tabular}{lccc}
\hline
Method & Buffer size & FAA ($\uparrow$) & FF ($\downarrow$) \\ \hline
ER & \multirow{6}{*}{1000} & 67.87\scriptsize{$\pm$0.57}   & 33.33\scriptsize{$\pm$1.28}  \\
BiC && 66.11\scriptsize{$\pm$1.76}   & 35.24\scriptsize{$\pm$1.64}  \\
GDumb && 67.14\scriptsize{$\pm$0.37}   & - \\
DER++&& 61.06\scriptsize{$\pm$0.87}   & 39.87\scriptsize{$\pm$0.99}  \\
Co$^2$L && 72.15\scriptsize{$\pm$1.32}   & 28.55\scriptsize{$\pm$1.56}  \\ 
DyTox && 77.61\scriptsize{$\pm$0.92}   & 8.26\scriptsize{$\pm$0.38}  \\
\hline
ER& \multirow{6}{*}{5000} & 82.53\scriptsize{$\pm$0.17}   & 16.46\scriptsize{$\pm$0.25}  \\
BiC&& 81.42\scriptsize{$\pm$0.85}   & 17.31\scriptsize{$\pm$1.02}  \\
GDumb&& 81.67\scriptsize{$\pm$0.02}   & -\\
DER++&& 83.94\scriptsize{$\pm$0.34}   & 14.55\scriptsize{$\pm$0.73}  \\
Co$^2$L&& 82.49\scriptsize{$\pm$0.89}   & 17.48\scriptsize{$\pm$1.80}  \\ 
DyTox && 88.15\scriptsize{$\pm$0.28}   & 3.64\scriptsize{$\pm$0.19}  \\
\hline
FT-seq& \multirow{5}{*}{0}    & 33.61\scriptsize{$\pm$0.85}   & 86.87\scriptsize{$\pm$0.20}  \\
EWC&& 47.01\scriptsize{$\pm$0.29}   & 33.27\scriptsize{$\pm$1.17}  \\
LwF&& 60.69\scriptsize{$\pm$0.63}   & 27.77\scriptsize{$\pm$2.17}  \\
L2P&& 83.86\scriptsize{$\pm$0.28} & 7.35\scriptsize{$\pm$0.38} \\
ESN && \bf{86.34\scriptsize{$\pm$0.52}} & \bf{4.76\scriptsize{$\pm$0.14}} \\
\hline
Upper-bound & -                     & 91.27\scriptsize{$\pm$0.18}  & - \\ \hline
\end{tabular}
\end{center}
\caption{Results on Split CIFAR-100 for class-incremental learning. \textbf{Bold}: best rehearsal-free results. 
All results except ESN, DyTox, and Upper-bound are copied from \cite{Wang_2022_CVPR}.
}
\label{table:cifar}
\end{table}

\noindent
\textbf{Comparison Methods.}
We compare ESN against the state-of-the-art CIL and DIL methods. 
Though we are a rehearsal-free incremental learning method, we also consider rehearsal-based methods that need the buffer to store exemplars for a more fair comparison.
Comparison methods are EWC~\cite{kirkpatrick2017overcoming}, LwF~\cite{li2017learning} ER~\cite{chaudhry2019tiny}, GDumb~\cite{prabhu2020gdumb}, BiC~\cite{wu2019large}, DER++~\cite{buzzega2020dark} and Co2L~\cite{cha2021co2l}, as well as the recently published transformer-based methods L2P~\cite{Wang_2022_CVPR} and DyTox~\cite{douillard2022dytox}.
To compare fairly, we use the same ViT models pre-trained on ImageNet (i.e., ViT-B/16~\cite{dosovitskiy2020vit}) for all the competitors as well as ESN. 
We use the joint training result as the upper-bound for ESN on all benchmarks.

\noindent
\textbf{Implementation details.}
We implement our method in PyTorch with two NVIDIA RTX 3090 GPUs.
The proposed ESN is insensitive to hyper-parameters.
We use the SGD optimizer and the cosine annealing learning rate scheduler with a initial learning rate of $0.01$ all benchmarks. 
We use $30$ epochs for Split CIFAR-100 and Split DomainNet, $10$ epochs for 5-Datasets and Core50.
We set the batch size of $128$ for all experiments.
Momentum and weight decay parameters are set to $0.9$ and $0.0005$, respectively. 
We use the ViT-B/16 pre-trained on ImageNet as backbone and the classifier is a class-attention block (CAB)~\cite{touvron2021going} with a linear projection.
The hyper-parameters of CAB is the same as ViT-B/16 except the MLP ratio is $0.5$, which has the parameters $3M$.
Due to the fact raw features extracted from pre-trained ViT are not suitable for all downstream tasks, we also add  parameters ($10\times768$) to the input, like~\cite{jia2022vpt}.
The candidate temperature set $\Psi$ is from a range of numbers from $0.001$ to $1.0$ with step of $0.001$.
We set the energy anchor $\Delta = -10$ and balance hyper-parameter $\lambda = 0.1$ for all benchmarks. 
Code will be available soon.

\begin{table}[]\small
\setlength{\tabcolsep}{10pt}
\begin{center}
\begin{tabular}{lccc}
\hline
Method      & Buffer size          & FAA ($\uparrow$)         & FF ($\downarrow$)    \\ \hline
ER & \multirow{5}{*}{250} & {80.32\scriptsize{$\pm$0.55}} & 15.69\scriptsize{$\pm$0.89}  \\
BiC && 78.74\scriptsize{$\pm$1.41}   & 21.15\scriptsize{$\pm$1.00}  \\ 
DER++     && 80.81\scriptsize{$\pm$0.07}   & 14.38\scriptsize{$\pm$0.35}  \\ 
Co$^2$L     && 82.25\scriptsize{$\pm$1.17}   & 17.52\scriptsize{$\pm$1.35}  \\
\hline
ER          & \multirow{5}{*}{500} & 84.26\scriptsize{$\pm$0.84} & 12.85\scriptsize{$\pm$0.62}  \\
BiC         && 85.53\scriptsize{$\pm$2.06}     & 10.27\scriptsize{$\pm$1.32}  \\
DER++       && 84.88\scriptsize{$\pm$0.57}   & 10.46\scriptsize{$\pm$1.02}  \\
Co$^2$L     && 86.05\scriptsize{$\pm$1.03}   & 12.28\scriptsize{$\pm$1.44}  \\ 
\hline
FT-seq      & \multirow{5}{*}{0}   & 20.12\scriptsize{$\pm$0.42}   & 94.63\scriptsize{$\pm$0.68}  \\
EWC && 50.93\scriptsize{$\pm$0.09}   & 34.94\scriptsize{$\pm$0.07}  \\
LwF && 47.91\scriptsize{$\pm$0.33}   & 38.01\scriptsize{$\pm$0.28}  \\
{L2P} && 81.14\scriptsize{$\pm$0.93}   & {4.64\scriptsize{$\pm$0.52}} \\
ESN && \bf{85.71\scriptsize{$\pm$1.47}}   & \bf{2.85\scriptsize{$\pm$0.61}}   \\ \hline
Upper-bound & -                    & 94.39\scriptsize{$\pm$0.21}   & -                            \\ \hline
\end{tabular}
\end{center}
\caption{Results on 5-Datasets for class-incremental learning. \textbf{Bold}: best rehearsal-free results. All results except ESN and Upper-bound are copied from \cite{Wang_2022_CVPR}.}
\label{table:5dataset}
\end{table}

\subsection{Comparison Results}
We compare the proposed ESN with the state-of-the-arts on Split CIFAR-100, Split DomainNet, 5-Datasets and CORe50. 
We run ESN for $5$ times with different random seeds and report the average results. 
\textbf{For fair comparison, all methods start from the same ImageNet pre-trained ViT-B/16.}

\noindent
\textbf{Results on Class-incremental learning benchmarks.}
% For Class-incremental learning, we conduct experiments on two incremental learning benchmarks: Split CIFAR-100 and 5-Datasets.
Table~\ref{table:cifar} and Table~\ref{table:5dataset} summarize the results on Split CIFAR-100 and 5-Datasets benchmarks respectively.
% It can be seen that our proposed method outperforms other counterparts by a large margin in terms of both accuracy and forgetting.
ESN achieves state-of-the-art performance without any rehearsal buffer in terms of average accuracy and forgetting.
We compute that ESN obtains a considerable relative improvement (an average of roughly $3.5\%$) over the best rehearsal-free methods.
We can see that most rehearsal-based methods significantly improve by storing more data.
That shows that rehearsal-based methods' performances highly depend on buffer size.
The outstanding performance of ESN indicates that the proposed anchor-based energy self-normalization can successfully aggregate all stage classifiers impartially.
And thus can get outstanding performance even without rehearsal buffer.

% Besides, they gain relatively little improvements on 5-Datasets.
% That is probably due to the vast task diversity of 5-Datasets leads to little profit from storing more samples.

% For rehearsal-free methods, EWC and LwF are two representative baseline methods which use regularization strategies to guide the parameter update. 
% L2P use prompt fine-tuning for pre-trained models 

\begin{table}[]\small
\setlength{\tabcolsep}{20pt}
\begin{center}
\begin{tabular}{lcc}
\hline
Method & Buffer size & FAA ($\uparrow$) \\ \hline
ER & \multirow{6}{*}{50/class} & 80.10\scriptsize{$\pm$0.56} \\
GDumb&& 74.92\scriptsize{$\pm$0.25} \\
BiC&& 79.28\scriptsize{$\pm$0.30} \\
DER++&& 79.70\scriptsize{$\pm$0.44} \\
Co$^2$L&& 79.75\scriptsize{$\pm$0.84}  \\
DyTox && 79.21\scriptsize{$\pm$0.10} \\ 
L2P&& 81.07\scriptsize{$\pm$0.13} \\ \hline
EWC & \multirow{4}{*}{0} & 74.82\scriptsize{$\pm$0.60}  \\
LwF && 75.45\scriptsize{$\pm$0.40}  \\
L2P && 78.33\scriptsize{$\pm$0.06}  \\
ESN & & \bf{91.80\scriptsize{$\pm$0.31}} \\ \hline
Upper-bound & - & 92.50\scriptsize{$\pm$0.11}      \\ \hline
\end{tabular}
\end{center}
\caption{Results on CORe50 for domain-incremental learning, in terms of final test accuracy. \textbf{Bold}: best rehearsal-free results. All results except ESN, DyTox, and Upper-bound are copied from \cite{Wang_2022_CVPR}.} 
\label{table:core50}
\end{table}

\noindent
\textbf{Results on Domain-incremental learning benchmarks.}
Table~\ref{table:core50} summarizes the results on the CORe50 benchmark. 
CORe50 is a challenging DIL benchmark that uses $8$ domains as train set and $3$ domains as test set.
That means test images do not belong to any training domains, and this benchmark mainly tests the generalization ability after incremental learning.
% the generalization ability of the model aimed to be improved with incremental learning process.
% and 
% This benchmark has $50$ classes and $11$ domains, of which $8$ domains are used as training data, and the rest of domains are regarded as a test set.
ESN achieves the best performance compared with other methods (about $17\%$ improvements over L2P) with the same ViT-B/16 pre-trained backbone.
Since there is no correct stage-ID for test images (no domain overlap), the accuracy of ESN comes from the ensemble voting strategy. 
% Let all classifiers trained so far do voting, and the robustness of predictions can be significantly improved.

% Because test images are all out-of-distribution data, the remarkable accuracy of our method mainly comes from our ensemble strategy.

\noindent
\textbf{Results on Cross-Domain Class-incremental learning benchmark.}
Cross-Domain Class-incremental learning is a more challenging scenario than traditional CIL settings.
% That's mainly due to the extensive large domain distribution between each stage leads to forgetting.
As shown in Table~\ref{table:domainnet}, ESN out-performs all other rehearsal-free methods a large margin (about $50\%$ improvement).
We can see that most class incremental learning algorithms fail to prevent catastrophic forgetting in the cross-domain setting to a great extent, as indicated by high final forgetting (FF) shown in Table~\ref{table:domainnet}.
Specially, some regularization-based methods, LwF and EWC, even perform worse than simply finetuning.
That is probably due to some regularization are not robust to large domain shift.
Our stage isolation learning strategy can preserve old knowledge successfully.
And the proposed anchor-based energy self-normalization strategy is robust to handle this challenging scenario.

\begin{table}[] \small
\setlength{\tabcolsep}{10pt}
\begin{center}
\begin{tabular}{lccc}
\hline
Method & Buffer size & FAA ($\uparrow$) & \multicolumn{1}{l}{FF ($\downarrow$)} \\ \hline
ER & \multirow{4}{*}{250} & {64.54\scriptsize{$\pm$1.06}} & 28.21\scriptsize{$\pm$0.45}  \\
BiC && 66.99\scriptsize{$\pm$1.27}& 19.91\scriptsize{$\pm$0.23}  \\ 
DER++ && 70.18\scriptsize{$\pm$0.37}& 21.31\scriptsize{$\pm$0.55}  \\ 
DyTox && 77.16\scriptsize{$\pm$0.72}   & 6.88\scriptsize{$\pm$0.31}  \\

\hline
ER & \multirow{4}{*}{500} & 70.90\scriptsize{$\pm$1.35} & 21.49\scriptsize{$\pm$0.61}  \\
BiC & & 68.19\scriptsize{$\pm$1.22} & 21.76\scriptsize{$\pm$0.39}  \\
DER++ & & 74.61\scriptsize{$\pm$0.27}   & 16.65\scriptsize{$\pm$0.94}  \\ 
DyTox && 79.6\scriptsize{$\pm$0.91}   & 5.87\scriptsize{$\pm$0.20}  \\
\hline

Finetune & \multirow{5}{*}{0} & 35.66\scriptsize{$\pm$2.73} & 59.89\scriptsize{$\pm$2.05} \\
EWC && 22.35\scriptsize{$\pm$1.86} & 76.11\scriptsize{$\pm$1.28} \\
LwF && 28.86\scriptsize{$\pm$1.92} & 64.91\scriptsize{$\pm$1.01} \\
L2P && 45.65\scriptsize{$\pm$0.82} & 15.26\scriptsize{$\pm$0.51} \\
ESN & &   \bf{68.76\scriptsize{$\pm$0.12}} & \bf{5.75\scriptsize{$\pm$0.23}}\\ \hline
Upper-bound & - & 82.53\scriptsize{$\pm$0.44} & - \\ \hline
\end{tabular}

\end{center}
\caption{Results on Split DomainNet for cross-domain class-incremental learning. \textbf{Bold}: best rehearsal-free results.
}
\label{table:domainnet}
\end{table}

\subsection{Ablation Study} 
\label{sec:abonarchi}

\noindent
\textbf{The effect of related components.}
To further study the effectiveness of ESN, we study the effect of our main components in Table~\ref{tab:ablation}.
Table~\ref{tab:ablation} (row 1) removes the proposed anchor-based energy self-normalization strategy $\mathcal{L}_{al}^{s}$, and keeps the other parts the same. The performance has a significant drop, suggesting that aligning all isolated classifiers to the same energy plane is the key issue in aggregating them impartially for final prediction.
Table~\ref{tab:ablation} (row 2) removes our proposed temperature selection strategy, and just using the default temperature $1$ without voting for prediction.
The results is slightly lower than ESN. 
The decrease suggests that using the proposed temperature calibration can further boost the performance.
% that though we have align all classifiers to the same energy plane, they still have slightly bias or difference.
% And Table~\ref{tab:ablation} (row 3) shows that using temperatures randomly selected from the candidate temperature pool $\$
Table~\ref{tab:ablation} (row 3) shares the same class-attention block (CAB) across tasks.
As the result shows, parameter isolation is important in tackling catastrophic forgetting and maintaining performance.

\begin{table}[] \small
\setlength{\tabcolsep}{13pt}
\begin{center}
\begin{tabular}{lll}
\hline
Ablated components          & FAA ($\uparrow$)   & FF ($\downarrow$)   \\ \hline
w/o energy self-normalization & 80.21 & 9.35 \\
w/o temperature calibration & 85.73 & 4.88 \\
% w/o temperature random select & 86.08 & 4.78 \\
w/o parameter isolation     & 83.94 & 6.42 \\ \hline
None                        & 86.34 & 4.76 \\ \hline
\end{tabular}
\caption{Ablation studies of the effect of related components. The experiments are performed on Split CIFAR-100. }
\label{tab:ablation}
\end{center}
\end{table}

\begin{table}[]\small
\setlength{\tabcolsep}{4pt}
\begin{center}
\begin{tabular}{lllllll}
\hline
Energy Anchor $\Delta$ & 0     & -1    & -3    & -5    & -10   & -15   \\ \hline
FAA ($\uparrow$)    & 85.96 & 85.60 & 85.59 & 86.20 & 86.34 & 86.25 \\
FF ($\downarrow$) & 5.08  & 5.21  & 5.56  & 4.59  & 4.76  & 4.98  \\ \hline
\end{tabular}
\caption{The effect of the energy anchor $\Delta$. The experiments are performed on Split CIFAR-100. 
}
\label{tab:effect_of_delta}
\end{center}
\end{table}

\noindent
\textbf{The effect of different $\Delta$.} 
% and $\lambda$
% $\Delta$ and $\lambda$ are two main
$\Delta$ is the main hyper-parameter of our proposed energy self-normalization loss, and we conduct an ablation study to investigate its effect.
Table~\ref{tab:effect_of_delta} shows the final results (FAA and FF) is insensitive to the value of $\Delta$.
That is probably because the most important thing is to normalize all classifiers to the same energy plane.
% And we do not need much effort to find the optimal value of that plane.
% Table~\ref{tab:effect_of_lambda} shows the effect of $\lambda$.
% When $\lambda$ is $0$, the energy alignment loss will have no effect.
% However, when $\lambda$ is $0.3$, the training process will collapse.
% We can see the energy alignment loss has positive effect to the final performance. 

% \begin{table}[] \small
% \setlength{\tabcolsep}{8pt}
% \begin{center}
% \begin{tabular}{lllll}
% \hline
% $\lambda$ & 0     & 0.1   & 0.2   & 0.3 \\ \hline
% FAA ($\%$) & 80.61 & 86.23 & 85.82 & NA  \\
% FF         & 12.31  & 4.59  & 4.84  & NA  \\ \hline
% \end{tabular}
% \caption{Impact of $\lambda$ Loss}
% \label{tab:effect_of_lambda}
% \end{center}
% \end{table}

\noindent
\textbf{The effect of different network architectures.}
In the main experiments, we mainly attach a class-attention block as a decoder to the pre-trained backbone.
We point out that other network architectures can also use our proposed energy self-normalization method.
Table~\ref{tab:ab_architectures} summarizes the results of using different architectures.
Here, we add two parameter isolation methods to demonstrate our idea: VPT~\cite{jia2022vpt} and DER~\cite{yan2021dynamically}.
VPT uses a small amount of task-specific learnable parameters into the input while freezing the other parts of the model to tune a pre-trained model to downstream tasks.
DER expands a new network for each new coming task.
The network can be any type, and we use both ResNet50 and ViT-B/16 for experiments.
We report the amount of expansion parameters for a single incremental stage in the Table~\ref{tab:ab_architectures}.
Though the amount of expansion parameters of VPT is significantly less than CAB, VPT needs almost ten times inference time than CAB.
That is because CAB works as a stage-specific decoder and uses a shared backbone to extract image features, which can decrease the computational expense.
DER-like methods have the same inference speed problem and perform worse than VPT and CAB. 
The worse performance of DER-like methods is probably because training large models on a small subset of a dataset has severe over-fitting.
% and we find 

% The most important thing is to train independent networks for every stage.
% We clarify here that our method does not depend on any network structure at all, and the architecture used in the implementation is to make our method as clear as possible.

\begin{table}[] \small
\setlength{\tabcolsep}{6pt}
\begin{center}
\begin{tabular}{lllll}
\hline
\multirow{2}{*}{Method} & \multicolumn{2}{l}{Expansion Parameters} & \multirow{2}{*}{FAA ($\uparrow$)} & \multirow{2}{*}{FF ($\downarrow$)} \\
& M & Relative increase ($\%$) & & \\ \hline
DER-ViT & 86.6 & 100 & 83.43 & 5.52 \\
DER-ResNet50 & 25.3 & 100 & 80.37 & 9.2 \\
VPT & 0.2 & 0.2 & 85.55 & 4.98 \\
CAB & 3.0 & 3.4 & 86.34 & 4.76 \\ \hline
\end{tabular}
\caption{Ablation studies of different network architectures. The experiments are performed on Split CIFAR-100. 
}
\label{tab:ab_architectures}
\end{center}
\end{table}

% \begin{table}[] \small
% \setlength{\tabcolsep}{6pt}
% \begin{center}
% \begin{tabular}{llll}
% \hline
% Method       & Expansion Parameters  & FAA ($\uparrow$) & FF ($\downarrow$) \\ \hline
% DER-ViT      & 86.6M      &  83.43   &  5.52  \\
% DER-ResNet50 & 25.3M        & 80.37  & 9.2  \\
% VPT & 1.8K & 85.55 & 4.98 \\
% CAB & 3.0M & 86.34 & 4.76 \\
% \hline
% \end{tabular}
% \caption{Ablation studies of different architectures. The experiments are performed on Split CIFAR-100. 
% }
% \label{tab:ab_architectures}
% \end{center}
% \end{table}

\section{Conclusion}

This paper proposes a novel a rehearsal-free stage-isolation based general incremental learning framework.
The proposed ESN learns stage-isolation classifiers for each stage, and uses then anchor-based energy self-normalization strategy to aggregate multiple isolated classifiers in an impartial way.
Furthermore, we propose a control parameter (temperature) finding method and propose a voting based inference augmentation strategy for robust inference.
% ESN is insensitive to hyper-parameters and can apply to wide range of network architectures.
Our experiments show that our method outperforms the current state-of-the-art on four large benchmarks \XP{by} a large margin and can handle general incremental learning scenarios.
 
\section*{Acknowledgements}
This work is funded by the National Key Research and Development Project of China (2019YFB1312000), the National Natural Science Foundation of China (62076195, 62206271, and U20B2052), the Fundamental Research Funds for the Central Universities (AUGA5710011522), and the Guangdong Basic and Applied Basic Research Foundation (2020B1515130004). This work is also supported by the Singapore Ministry of Education (MOE) Academic Research Fund (AcRF) Tier 1 grant (MSS21C002).
\bibliography{bib_OOD-IL}

\begin{thebibliography}{44}
\providecommand{\natexlab}[1]{#1}

\bibitem[{Al~Rahhal et~al.(2022)Al~Rahhal, Bazi, Al-Dayil, Alwadei, Ammour, and
  Alajlan}]{al2022energy}
Al~Rahhal, M.~M.; Bazi, Y.; Al-Dayil, R.; Alwadei, B.~M.; Ammour, N.; and
  Alajlan, N. 2022.
\newblock Energy-based learning for open-set classification in remote sensing
  imagery.
\newblock \emph{International Journal of Remote Sensing}, 1--11.

\bibitem[{Buzzega et~al.(2020)Buzzega, Boschini, Porrello, Abati, and
  Calderara}]{buzzega2020dark}
Buzzega, P.; Boschini, M.; Porrello, A.; Abati, D.; and Calderara, S. 2020.
\newblock Dark experience for general continual learning: a strong, simple
  baseline.
\newblock \emph{NeurIPS}.

\bibitem[{Cha, Lee, and Shin(2021)}]{cha2021co2l}
Cha, H.; Lee, J.; and Shin, J. 2021.
\newblock Co2l: Contrastive continual learning.
\newblock In \emph{ICCV}.

\bibitem[{Chaudhry et~al.(2019)Chaudhry, Rohrbach, Elhoseiny, Ajanthan,
  Dokania, Torr, and Ranzato}]{chaudhry2019tiny}
Chaudhry, A.; Rohrbach, M.; Elhoseiny, M.; Ajanthan, T.; Dokania, P.~K.; Torr,
  P.~H.; and Ranzato, M. 2019.
\newblock On tiny episodic memories in continual learning.
\newblock \emph{arXiv preprint arXiv:1902.10486}.

\bibitem[{De~Lange et~al.(2021)De~Lange, Aljundi, Masana, Parisot, Jia,
  Leonardis, Slabaugh, and Tuytelaars}]{de2021continual}
De~Lange, M.; Aljundi, R.; Masana, M.; Parisot, S.; Jia, X.; Leonardis, A.;
  Slabaugh, G.; and Tuytelaars, T. 2021.
\newblock A continual learning survey: Defying forgetting in classification
  tasks.
\newblock \emph{IEEE transactions on pattern analysis and machine
  intelligence}, 44(7): 3366--3385.

\bibitem[{Dong et~al.(2021)Dong, Hong, Tao, Chang, Wei, and Gong}]{dong2021few}
Dong, S.; Hong, X.; Tao, X.; Chang, X.; Wei, X.; and Gong, Y. 2021.
\newblock Few-shot class-incremental learning via relation knowledge
  distillation.
\newblock In \emph{Proceedings of the AAAI Conference on Artificial
  Intelligence}, volume~35, 1255--1263.

\bibitem[{Dosovitskiy et~al.(2021)Dosovitskiy, Beyer, Kolesnikov, Weissenborn,
  Zhai, Unterthiner, Dehghani, Minderer, Heigold, Gelly, Uszkoreit, and
  Houlsby}]{dosovitskiy2020vit}
Dosovitskiy, A.; Beyer, L.; Kolesnikov, A.; Weissenborn, D.; Zhai, X.;
  Unterthiner, T.; Dehghani, M.; Minderer, M.; Heigold, G.; Gelly, S.;
  Uszkoreit, J.; and Houlsby, N. 2021.
\newblock An Image is Worth 16x16 Words: Transformers for Image Recognition at
  Scale.
\newblock \emph{ICLR}.

\bibitem[{Douillard et~al.(2022)Douillard, Ram{\'e}, Couairon, and
  Cord}]{douillard2022dytox}
Douillard, A.; Ram{\'e}, A.; Couairon, G.; and Cord, M. 2022.
\newblock Dytox: Transformers for continual learning with dynamic token
  expansion.
\newblock In \emph{Proceedings of the IEEE/CVF Conference on Computer Vision
  and Pattern Recognition}, 9285--9295.

\bibitem[{Du and Mordatch(2019)}]{du2019implicit}
Du, Y.; and Mordatch, I. 2019.
\newblock Implicit generation and modeling with energy based models.
\newblock \emph{Advances in Neural Information Processing Systems}, 32.

\bibitem[{Ebrahimi et~al.(2020)Ebrahimi, Meier, Calandra, Darrell, and
  Rohrbach}]{ebrahimi2020adversarial}
Ebrahimi, S.; Meier, F.; Calandra, R.; Darrell, T.; and Rohrbach, M. 2020.
\newblock Adversarial continual learning.
\newblock In \emph{European Conference on Computer Vision}, 386--402. Springer.

\bibitem[{Fort, Ren, and Lakshminarayanan(2021)}]{fort2021exploring}
Fort, S.; Ren, J.; and Lakshminarayanan, B. 2021.
\newblock Exploring the limits of out-of-distribution detection.
\newblock \emph{Advances in Neural Information Processing Systems}, 34:
  7068--7081.

\bibitem[{Grathwohl et~al.(2019)Grathwohl, Wang, Jacobsen, Duvenaud, Norouzi,
  and Swersky}]{grathwohl2019your}
Grathwohl, W.; Wang, K.-C.; Jacobsen, J.-H.; Duvenaud, D.; Norouzi, M.; and
  Swersky, K. 2019.
\newblock Your classifier is secretly an energy based model and you should
  treat it like one.
\newblock \emph{arXiv preprint arXiv:1912.03263}.

\bibitem[{Hendrycks et~al.(2019)Hendrycks, Basart, Mazeika, Mostajabi,
  Steinhardt, and Song}]{hendrycks2019scaling}
Hendrycks, D.; Basart, S.; Mazeika, M.; Mostajabi, M.; Steinhardt, J.; and
  Song, D. 2019.
\newblock Scaling out-of-distribution detection for real-world settings.
\newblock \emph{arXiv preprint arXiv:1911.11132}.

\bibitem[{Hou et~al.(2019)Hou, Pan, Loy, Wang, and Lin}]{hou2019learning}
Hou, S.; Pan, X.; Loy, C.~C.; Wang, Z.; and Lin, D. 2019.
\newblock Learning a unified classifier incrementally via rebalancing.
\newblock In \emph{Proceedings of the IEEE/CVF Conference on Computer Vision
  and Pattern Recognition}, 831--839.

\bibitem[{Jia et~al.(2022)Jia, Tang, Chen, Cardie, Belongie, Hariharan, and
  Lim}]{jia2022vpt}
Jia, M.; Tang, L.; Chen, B.-C.; Cardie, C.; Belongie, S.; Hariharan, B.; and
  Lim, S.-N. 2022.
\newblock Visual Prompt Tuning.
\newblock In \emph{European Conference on Computer Vision (ECCV)}.

\bibitem[{Joseph et~al.(2022)Joseph, Khan, Khan, Anwer, and
  Balasubramanian}]{joseph2022energy}
Joseph, K.; Khan, S.; Khan, F.~S.; Anwer, R.~M.; and Balasubramanian, V.~N.
  2022.
\newblock Energy-based Latent Aligner for Incremental Learning.
\newblock In \emph{Proceedings of the IEEE/CVF Conference on Computer Vision
  and Pattern Recognition}, 7452--7461.

\bibitem[{Kirkpatrick et~al.(2017)Kirkpatrick, Pascanu, Rabinowitz, Veness,
  Desjardins, Rusu, Milan, Quan, Ramalho, Grabska-Barwinska
  et~al.}]{kirkpatrick2017overcoming}
Kirkpatrick, J.; Pascanu, R.; Rabinowitz, N.; Veness, J.; Desjardins, G.; Rusu,
  A.~A.; Milan, K.; Quan, J.; Ramalho, T.; Grabska-Barwinska, A.; et~al. 2017.
\newblock Overcoming catastrophic forgetting in neural networks.
\newblock \emph{Proceedings of the national academy of sciences}, 114(13):
  3521--3526.

\bibitem[{Knoblauch, Husain, and Diethe(2020)}]{knoblauch2020optimal}
Knoblauch, J.; Husain, H.; and Diethe, T. 2020.
\newblock Optimal continual learning has perfect memory and is np-hard.
\newblock In \emph{ICML}.

\bibitem[{Krizhevsky, Hinton et~al.(2009)}]{krizhevsky2009learning}
Krizhevsky, A.; Hinton, G.; et~al. 2009.
\newblock Learning multiple layers of features from tiny images.

\bibitem[{LeCun et~al.(2006)LeCun, Chopra, Hadsell, Ranzato, and
  Huang}]{lecun2006tutorial}
LeCun, Y.; Chopra, S.; Hadsell, R.; Ranzato, M.; and Huang, F. 2006.
\newblock A tutorial on energy-based learning.
\newblock \emph{Predicting structured data}, 1(0).

\bibitem[{Li et~al.(2019)Li, Zhou, Wu, Socher, and Xiong}]{li2019learn}
Li, X.; Zhou, Y.; Wu, T.; Socher, R.; and Xiong, C. 2019.
\newblock Learn to grow: A continual structure learning framework for
  overcoming catastrophic forgetting.
\newblock In \emph{International Conference on Machine Learning}, 3925--3934.
  PMLR.

\bibitem[{Li and Hoiem(2017)}]{li2017learning}
Li, Z.; and Hoiem, D. 2017.
\newblock Learning without forgetting.
\newblock \emph{IEEE transactions on pattern analysis and machine
  intelligence}, 40(12): 2935--2947.

\bibitem[{Liu et~al.(2020)Liu, Wang, Owens, and Li}]{liu2020energy}
Liu, W.; Wang, X.; Owens, J.; and Li, Y. 2020.
\newblock Energy-based out-of-distribution detection.
\newblock \emph{Advances in Neural Information Processing Systems}, 33:
  21464--21475.

\bibitem[{Liu et~al.(2022)Liu, Hong, Tao, Dong, Shi, and Gong}]{liu2022model}
Liu, Y.; Hong, X.; Tao, X.; Dong, S.; Shi, J.; and Gong, Y. 2022.
\newblock Model Behavior Preserving for Class-Incremental Learning.
\newblock \emph{IEEE Transactions on Neural Networks and Learning Systems}.

\bibitem[{Lomonaco and Maltoni(2017)}]{lomonaco2017core50}
Lomonaco, V.; and Maltoni, D. 2017.
\newblock Core50: a new dataset and benchmark for continuous object
  recognition.
\newblock In \emph{Conference on Robot Learning}, 17--26. PMLR.

\bibitem[{Mai et~al.(2022)Mai, Li, Jeong, Quispe, Kim, and
  Sanner}]{mai2022online}
Mai, Z.; Li, R.; Jeong, J.; Quispe, D.; Kim, H.; and Sanner, S. 2022.
\newblock Online continual learning in image classification: An empirical
  survey.
\newblock \emph{Neurocomputing}, 469: 28--51.

\bibitem[{Peng et~al.(2019)Peng, Bai, Xia, Huang, Saenko, and
  Wang}]{peng2019moment}
Peng, X.; Bai, Q.; Xia, X.; Huang, Z.; Saenko, K.; and Wang, B. 2019.
\newblock Moment matching for multi-source domain adaptation.
\newblock In \emph{Proceedings of the IEEE International Conference on Computer
  Vision}, 1406--1415.

\bibitem[{Prabhu, Torr, and Dokania(2020)}]{prabhu2020gdumb}
Prabhu, A.; Torr, P.~H.; and Dokania, P.~K. 2020.
\newblock Gdumb: A simple approach that questions our progress in continual
  learning.
\newblock In \emph{ECCV}.

\bibitem[{Riemer et~al.(2018)Riemer, Cases, Ajemian, Liu, Rish, Tu, and
  Tesauro}]{riemer2018learning}
Riemer, M.; Cases, I.; Ajemian, R.; Liu, M.; Rish, I.; Tu, Y.; and Tesauro, G.
  2018.
\newblock Learning to Learn without Forgetting by Maximizing Transfer and
  Minimizing Interference.
\newblock In \emph{ICLR}.

\bibitem[{Serra et~al.(2018)Serra, Suris, Miron, and
  Karatzoglou}]{serra2018overcoming}
Serra, J.; Suris, D.; Miron, M.; and Karatzoglou, A. 2018.
\newblock Overcoming catastrophic forgetting with hard attention to the task.
\newblock In \emph{International Conference on Machine Learning}, 4548--4557.
  PMLR.

\bibitem[{Shin et~al.(2017)Shin, Lee, Kim, and Kim}]{shin2017continual}
Shin, H.; Lee, J.~K.; Kim, J.; and Kim, J. 2017.
\newblock Continual learning with deep generative replay.
\newblock \emph{Advances in neural information processing systems}, 30.

\bibitem[{Tang et~al.(2021)Tang, Miao, Peng, Wu, Shi, Gu, Tian, and
  Wang}]{tang2021codes}
Tang, K.; Miao, D.; Peng, W.; Wu, J.; Shi, Y.; Gu, Z.; Tian, Z.; and Wang, W.
  2021.
\newblock CODEs: Chamfer Out-of-Distribution Examples against Overconfidence
  Issue.
\newblock In \emph{Proceedings of the IEEE/CVF International Conference on
  Computer Vision}, 1153--1162.

\bibitem[{Tao et~al.(2020)Tao, Chang, Hong, Wei, and Gong}]{tao2020topology}
Tao, X.; Chang, X.; Hong, X.; Wei, X.; and Gong, Y. 2020.
\newblock Topology-preserving class-incremental learning.
\newblock In \emph{European Conference on Computer Vision}, 254--270. Springer.

\bibitem[{Touvron et~al.(2021)Touvron, Cord, Sablayrolles, Synnaeve, and
  J{\'e}gou}]{touvron2021going}
Touvron, H.; Cord, M.; Sablayrolles, A.; Synnaeve, G.; and J{\'e}gou, H. 2021.
\newblock Going deeper with image transformers.
\newblock In \emph{Proceedings of the IEEE/CVF International Conference on
  Computer Vision}, 32--42.

\bibitem[{Wang et~al.(2022{\natexlab{a}})Wang, Li, Feng, and
  Zhang}]{wang2022vim}
Wang, H.; Li, Z.; Feng, L.; and Zhang, W. 2022{\natexlab{a}}.
\newblock ViM: Out-Of-Distribution with Virtual-logit Matching.
\newblock In \emph{Proceedings of the IEEE/CVF Conference on Computer Vision
  and Pattern Recognition}, 4921--4930.

\bibitem[{Wang, Huang, and Hong(2022)}]{wang2022s}
Wang, Y.; Huang, Z.; and Hong, X. 2022.
\newblock S-Prompts Learning with Pre-trained Transformers: An Occam's Razor
  for Domain Incremental Learning.
\newblock In \emph{Conference on Neural Information Processing Systems
  (NeurIPS)}.

\bibitem[{Wang et~al.(2023)Wang, Ma, Huang, Wang, Su, and
  Hong}]{wang2023isolation}
Wang, Y.; Ma, Z.; Huang, Z.; Wang, Y.; Su, Z.; and Hong, X. 2023.
\newblock Isolation and Impartial Aggregation: A Paradigm of Incremental
  Learning without Interference.
\newblock In \emph{the Proceedings of the 37th AAAI Conference on Artificial
  Intelligence (AAAI 2023)}.

\bibitem[{Wang et~al.(2022{\natexlab{b}})Wang, Zhang, Lee, Zhang, Sun, Ren, Su,
  Perot, Dy, and Pfister}]{Wang_2022_CVPR}
Wang, Z.; Zhang, Z.; Lee, C.-Y.; Zhang, H.; Sun, R.; Ren, X.; Su, G.; Perot,
  V.; Dy, J.; and Pfister, T. 2022{\natexlab{b}}.
\newblock Learning To Prompt for Continual Learning.
\newblock In \emph{Proceedings of the IEEE/CVF Conference on Computer Vision
  and Pattern Recognition (CVPR)}, 139--149.

\bibitem[{Wu et~al.(2019)Wu, Chen, Wang, Ye, Liu, Guo, and Fu}]{wu2019large}
Wu, Y.; Chen, Y.; Wang, L.; Ye, Y.; Liu, Z.; Guo, Y.; and Fu, Y. 2019.
\newblock Large scale incremental learning.
\newblock In \emph{CVPR}.

\bibitem[{Xie, Yan, and He(2022)}]{xie2022general}
Xie, J.; Yan, S.; and He, X. 2022.
\newblock General Incremental Learning with Domain-aware Categorical
  Representations.
\newblock In \emph{Proceedings of the IEEE/CVF Conference on Computer Vision
  and Pattern Recognition}, 14351--14360.

\bibitem[{Xu and Zhang(2020)}]{xu2020aanet}
Xu, H.; and Zhang, J. 2020.
\newblock Aanet: Adaptive aggregation network for efficient stereo matching.
\newblock In \emph{Proceedings of the IEEE/CVF Conference on Computer Vision
  and Pattern Recognition}, 1959--1968.

\bibitem[{Yan, Xie, and He(2021)}]{yan2021dynamically}
Yan, S.; Xie, J.; and He, X. 2021.
\newblock Der: Dynamically expandable representation for class incremental
  learning.
\newblock In \emph{Proceedings of the IEEE/CVF Conference on Computer Vision
  and Pattern Recognition}, 3014--3023.

\bibitem[{Zhang et~al.(2022)Zhang, Dong, Chen, Tian, Gong, and Hong}]{9932643}
Zhang, X.; Dong, S.; Chen, J.; Tian, Q.; Gong, Y.; and Hong, X. 2022.
\newblock Deep Class-Incremental Learning From Decentralized Data.
\newblock \emph{IEEE Transactions on Neural Networks and Learning Systems},
  1--14.

\bibitem[{Zhao et~al.(2022)Zhao, Chen, Xiao, Ju, and Xia}]{zhao2022energy}
Zhao, B.; Chen, C.; Xiao, X.; Ju, Q.; and Xia, S. 2022.
\newblock Energy Alignment for Bias Rectification in Class Incremental
  Learning.
\newblock In \emph{ICASSP 2022-2022 IEEE International Conference on Acoustics,
  Speech and Signal Processing (ICASSP)}, 3513--3517. IEEE.

\end{thebibliography}

\end{document}